\documentclass[11pt,a4paper]{article}
\usepackage[hyperref]{emnlp2020}
\usepackage{times}
\usepackage{latexsym}

\usepackage{microtype}

\aclfinalcopy %
\def\aclpaperid{522} %

\usepackage{xspace}
\usepackage{multirow}
\usepackage{hhline}
\usepackage{amsmath,amsfonts,amssymb}
\usepackage{pgfplots}
\pgfplotsset{compat=1.9}
\usepackage{graphicx}
\graphicspath{ {images/} }
\usepackage{xcolor}
\usepackage{float}
\restylefloat{table}
\usepackage{bm}
\usepackage{soul}
\usepackage[textsize=tiny]{todonotes}   %
\usepackage[inline]{enumitem}
\usepackage{textcomp}
\usepackage{url}

\newcommand{\rnn}{\textsc{rnn}\xspace}

\newcommand{\lm}{\textsc{lm}\xspace}

\newcommand{\tm}{\textsc{tm}\xspace}
\newcommand{\smt}{\textsc{smt}\xspace}

\newcommand{\bleu}{\textsc{bleu}\xspace}
\newcommand{\nmt}{\textsc{nmt}\xspace}
\newcommand{\ls}{\textsc{ls}\xspace}
\newcommand{\kd}{\textsc{kd}\xspace}

\newcommand{\tto}{$\rightarrow$}

\usepackage{lettrine}
\newcommand{\GrantNo}{825299}
\newcommand{\ProjectName}{Gourmet}
\newcommand{\ProjectType}{Research and Innovation Action}

\title{Language Model Prior for Low-Resource Neural Machine Translation}

 \author{Christos Baziotis, Barry Haddow \and Alexandra Birch\\
    Institute for Language, Cognition and Computation \\
    School of Informatics, University of Edinburgh\\
    10 Crichton Street, Edinburgh EH8 9AB \\
  \texttt{c.baziotis@ed.ac.uk}, \texttt{bhaddow@inf.ed.ac.uk}, \texttt{a.birch@ed.ac.uk}}

\date{}

\begin{document}
\maketitle
\begin{abstract}
The scarcity of large parallel corpora is an important obstacle for neural machine translation.
A common solution is to exploit the knowledge of language models (\lm) trained on abundant monolingual data.
In this work, we propose a novel approach to incorporate a \lm as prior in a neural translation model~(\tm).
Specifically, we add a regularization term, which pushes the output distributions of the \tm to be probable under the \lm prior, while avoiding wrong predictions when the \tm ``disagrees'' with the \lm.
This objective relates to knowledge distillation, where the \lm can be viewed as teaching the \tm about the target language.
The proposed approach does not compromise decoding speed, because the \lm is used only at training time, 
unlike previous work that requires it during inference.
We present an analysis on the effects that different methods have on the distributions of the \tm.
Results on two low-resource machine translation datasets show clear improvements even with limited monolingual data.
\end{abstract}

\newlist{contributions}{enumerate*}{1}
\setlist[contributions]{label=(\arabic*)}

\abovedisplayskip=6pt
\abovedisplayshortskip=6pt
\belowdisplayskip=6pt
\belowdisplayshortskip=6pt

\section{Introduction}
Neural machine translation (\nmt)~\cite{sutskever2014sequence, Bahdanau2014,vaswani2017attention} relies heavily on large parallel corpora~\cite{koehn-knowles-2017-six} and needs careful hyper-parameter tuning, in order to work in low-resource settings~\cite{sennrich-zhang-2019-revisiting}.
A popular approach for addressing data scarcity is to exploit abundant monolingual corpora via data augmentation techniques, such as back-translation~\cite{sennrich-etal-2016-improving}. 
Although back-translation usually leads to significant performance gains~\cite{hoang-etal-2018-iterative}, 
it requires training separate models and expensive translation of large amounts of monolingual data.
However, when faced with lack of training data, a more principled approach is to consider exploiting prior information.%
  
Language models (\lm) trained on target-side monolingual data have been used for years as priors in statistical machine translation (\smt)~\cite{brown-etal-1993-mathematics} via the noisy channel model. %
This approach has been adopted to \nmt, with the neural noisy channel~\cite{Yu2016TheNN, yee-etal-2019-simple}.
However, neural noisy channel models face a computational challenge, because they model the ``reverse translation probability'' $p(x|y)$.
Specifically, they require multiple passes over the source sentence $x$ as they generate the target sentence $y$, or sophisticated architectures to reduce the passes.

\lm\/s have also been used in \nmt for re-weighting the predictions of translation models (\tm), or as additional context, via \lm-fusion~\cite{gulcehre2015using, Sriram2018, stahlberg-etal-2018-simple}.
But, as the \lm is required during decoding, it adds a significant computation overhead.
Another challenge is balancing the \tm and the \lm, whose ratio is either fixed~\cite{stahlberg-etal-2018-simple} or requires changing the model architecture~\cite{gulcehre2015using, Sriram2018}.

In this work, we propose to use a \lm trained on target-side monolingual corpora as a weakly informative prior.
We add a regularization term, which drives the output distributions of the \tm to be probable under the distributions of the \lm.
This gives flexibility to the \tm, by enabling it to deviate from the \lm when needed, unlike fusion methods that change the decoder's distributions, which can introduce translation errors.
The \lm ``teaches'' the \tm about the target language similar to knowledge distillation~\cite{bucilua2006model, hinton2015distilling}.
This method works by simply changing the training objective and does not require any changes to the model architecture.
Importantly, the \lm is separated from the \tm, which means that it is needed only during training, therefore we can decode faster than fusion or neural noisy channel.
We also note that this method is not intended as a replacement to other techniques that use monolingual data, such as back-translation, but is orthogonal to them.

We make the following contributions:
\begin{enumerate}[topsep=1pt,itemsep=-0.8ex,partopsep=1ex,parsep=1ex]
    \item We propose a simple and principled way for incorporating prior information from \lm\/s in \nmt, by adding an extra regularization term (\S~\ref{sec:lm-prior}).
    Also, this approach enables fast decoding, by requiring the \lm only during training.
    \item We report promising results (\S~\ref{sec:results}) in two low-resource translation datasets. 
    We find that the proposed \lm-prior yields clear improvements even with limited monolingual data.
    \item We perform an analysis (\S~\ref{sec:analysis}) on the effects that different methods have on the output distributions of the \tm, and show how this can lead to translation errors.
\end{enumerate}
\section{Background}
\nmt models trained with maximum likelihood estimation, model directly the probability $p(\bm{y}|\bm{x})$ of the target sentence $\bm{y}$ given the source sentence $\bm{x}$:
\begin{align*}
    \hat{\bm{y}} &= \arg\max\limits_{\bm{y}} \log p(\bm{y}|\bm{x}) %
\end{align*}
Modeling directly $p(\bm{y}|\bm{x})$ requires large amounts of parallel sentences to learn a good model and \nmt lacks a principled way for leveraging monolingual data.
In this section we review approaches that exploit prior information encoded in \lm\/s or the signal from the language modeling task.

\paragraph{Noisy Channel Model}
\smt~\cite{Koehn:2010:SMT:1734086} employs  Bayes' rule which offers a natural way for exploiting monolingual data, using a target-side \lm based on the so called ``noisy channel'' model~\cite{shannon1949mathematical}. 
Instead of directly modeling $p(\bm{y}|\bm{x})$, it models the ``reverse translation probability'' $p(\bm{x}|\bm{y})$, by rewriting $p(\bm{y}|\bm{x}) \propto p(\bm{x}|\bm{y}) \times p(\bm{y})$.
It selects words that are both \textit{a priori} likely with $p(\bm{y}_i)$ and ``explain well'' the input with $p(\bm{x}|y_i)$. 
This idea has been adopted to \nmt with neural noisy channel, but it has two fundamental limitations.
First, during decoding the model has to alternate between generating the output and scoring the input~\cite{Yu2016TheNN, yu2019putting} or perform multiple forward passes~\cite{yee-etal-2019-simple} over $\bm{x}$.
And crucially, since the \lm is part of the network it has to  also be used during inference, which adds a computational constraint on its size.

\paragraph{Fusion}
~\citet{gulcehre2015using} proposed to incorporate pretrained \lm\/s in \nmt, using \textit{shallow-} and \textit{deep-fusion}. 
In \textit{shallow-fusion}, the \lm re-weights the \tm's scores via log-linear interpolation:
\begin{align*}
    \log p({\bm{y}_t}) = &(1-\beta) \,\, \log p_{\tm}(\bm{y}_t|\bm{y}_{<t}, \bm{x}) \nonumber \\ 
                         &+ \beta \,\, \log p_{\lm}(\bm{y}_t|\bm{y}_{<t}) %
\end{align*}
In \textit{deep fusion}, they alter the model architecture to include the hidden states of a \rnn-\lm~\cite{mikolov2011rnnlm} as additional features for predicting the next word in the decoder, which are weighted with a controller mechanism (i.e., gating).
In both approaches, the \tm and \lm are first trained independently and are combined later.
\citet{Sriram2018} extend these ideas with \textit{cold-fusion}, where they train the \tm from scratch with the \lm, using its logits as features, instead of its \lm hidden states.
\citet{stahlberg-etal-2018-simple} simplify this, by training a \tm together with a fixed \lm, using combinations of the \tm's and \lm's outputs.
By training the \tm with the assistance of the \lm, the motivation is that the \tm will rely on the \lm for fluency, whereas the \tm will be able to focus on modeling the source.
They report the best results with the \textsc{postnorm} method, 
outperforming other \lm-fusion techniques.
\textsc{postnorm} parameterizes $p(\bm{y}_t)$ as follows:
{
\thinmuskip=0mu\medmuskip=0mu\thickmuskip=0mu
\begin{align*}
p(\bm{y}_t)=\text{softmax}(\log p_{\tm}(\bm{y}_t|y_{<t},\bm{x})+\log  p_{\lm}(\bm{y}_t))
\end{align*}
}
It is practically the same as \textit{shallow-fusion}, but with the \lm used \textit{also} during training, instead of used just in inference, and interpolating with $\lambda$=1. 

Fusion methods face the same computational limitation as noisy channel, since the \lm needs to be used during inference.
Also, probability interpolation methods, such as shallow fusion or \textsc{postnorm}, use a fixed weight for all time-steps, which can lead to translation errors.
Gated fusion~\cite{gulcehre2015using, Sriram2018} is more flexible, but requires changing the network architecture.

\paragraph{Other Approaches}
Transfer-learning is another approach for exploiting pretrained \lm\/s. 
\citet{ramachandran-etal-2017-unsupervised}, first proposed to use \lm\/s trained on monolingual corpora to initialize the encoder and decoder of a \tm. \citet{skorokhodov-etal-2018-semi} extended this idea to Transformer architectures~\cite{vaswani2017attention}. 
This approach requires the \tm to have identical architecture to the \lm, which can be a limitation if the \lm is huge.

\citet{domhan-hieber-2017-using} used language modeling as extra signal, by training the decoder of a \tm also as a \lm on target-side monolingual data. \citet{sennrich-etal-2016-improving} replaced the source with a \textsc{null} token, while training on monolingual data. Both, reported mixed results, with marginal gains.

\section{Language Model Prior}\label{sec:lm-prior}
We propose to move the \lm out of the \tm and use it as a prior over its decoder,
by employing posterior regularization (\textsc{pr})~\cite{ganchev2010posterior}. 
\textsc{pr} incorporates prior information, by imposing soft constraints on a model's posterior distributions, 
which is much easier than putting Bayesian priors over all the parameters of a deep neural network. 
\begin{align}
    \mathcal{L} = 
              &\sum_{t=1}^{N} 
               - \log p_{\tm}(\bm{y}_t|\bm{y}_{<t}, \bm{x})  \label{eq:prior} \\
              &+ \lambda \, D_{\textsc{kl}}(
                  p_{\tm}(\bm{y}_t|\bm{y}_{<t}, \bm{x})
                   \parallel
                  p_{\lm}(\bm{y}_t|\bm{y}_{<t})
                  ) \nonumber 
\end{align}
The first term is the standard translation objective $\mathcal{L}_{\textsc{mt}}$ and the second is the regularization term  $\mathcal{L}_{\textsc{kl}}$, which we interpret as a weakly informative prior over the \tm's distributions $p_{\tm}$, that expresses partial information about $\bm{y}$.
$\mathcal{L}_{\textsc{kl}}$ is defined as the Kullback-Leibler divergence between the output distributions of the \tm and the \lm, weighted by $\lambda$.

This formulation gives flexibility to the model, unlike probability interpolation, such as in fusion methods.
For example, \textsc{postnorm} multiplies the probabilities of the \lm and \tm, which is the same as applying a logical \textsc{and} operation, where only words that are probable under \textit{both} distributions will receive non-negligible probabilities. 
This prevents the model from generating the correct word when there is a large ``disagreement'' between the \tm and the \lm, which is inevitable as the \lm is not aware of the source sentence (i.e., unconditional).
However, by using the \lm-prior we do \textit{not} change the outputs of the \tm.
$\mathcal{L}_{\textsc{kl}}$ pushes the \tm to stay \textit{on average} close to the prior, but crucially, it enables the \tm to deviate from it when needed, for example to copy words from the source.

Secondly, the \lm is no longer part of the network.
This means that we can do inference using only the \tm, unlike fusion or neural noisy channel, which require the \lm for both training and decoding.
By lifting this computational overhead, we enable the use of large pretrained models \lm\/s~(BERT;~\citet{devlin2019bert}, GPT-2;~\citet{radford2019language}), without compromising speed or efficiency.

\subsection{Relation to Knowledge Distillation}
The regularization term in Eq.~\eqref{eq:prior} resembles knowledge distillation (\kd)~\cite{ba2014deep, bucilua2006model, hinton2015distilling}, where the soft output probabilities of a big teacher model are used to train a small compact student model, by minimizing their $D_{\textsc{kl}}$. 
However, in standard \kd the teacher is trained on the \textit{same} task as the student, like in \kd for machine translation~\cite{kim-rush-2016-sequence}.
However, the proposed \lm-prior is trained on a different task that requires only monolingual data, unlike \tm teachers that require parallel data.

We exploit this connection to \kd and following~\newcite{hinton2015distilling} we use a softmax-temperature parameter $\tau\geq1$ to control the smoothness of the output distributions $p_i= \frac{\exp(s_i / \tau)}{\sum_{j}^{ }\exp(s_j/\tau))}$, where $s_i$ is the un-normalized score of each word $i$ (i.e., logit). 
Higher values of $\tau$ produce smoother distributions. 
Intuitively, this controls how much information encoded in the tail of the \lm's distributions, we expose to the \tm.
Specifically, a well trained \lm will generate distributions with high probability for a few words, leaving others with probabilities close to zero.
By increasing $\tau$ we expose extra information to the \tm, because we reveal more low-probability words that the \lm found similar to the predicted word.

We use $\tau>1$ \textit{only} for computing the $D_{\textsc{kl}}$ between the distributions of the \tm and the \lm and is the same for both.
The magnitude of $D_{\textsc{kl}}$ scales as $1/\tau^2$, so it is important to multiply its output with $\tau^2$ to keep the scale of the $\mathcal{L}_{\textsc{kl}}$  loss invariant to $\tau$.
Otherwise, this would implicitly change the weight to $\lambda$ applied to $\mathcal{L}_{\textsc{kl}}$.
Finally, we re-write the regularization term of Eq.~\eqref{eq:prior} as follows:
{
\thinmuskip=0mu\medmuskip=0mu\thickmuskip=0mu
\begin{align*}
    \mathcal{L}_{\textsc{kl}}=\bm{\tau}^2
    D_{\textsc{kl}}(
        p_{\tm}(\bm{y}_t|\bm{y}_{<t},\bm{x};\bm{\tau})
        \parallel 
        p_{\lm}(\bm{y}_t|\bm{y}_{<t};\bm{\tau}))
\end{align*}
}

\begin{figure}[t]
\centering
	\includegraphics[width=0.9\columnwidth, page=6]{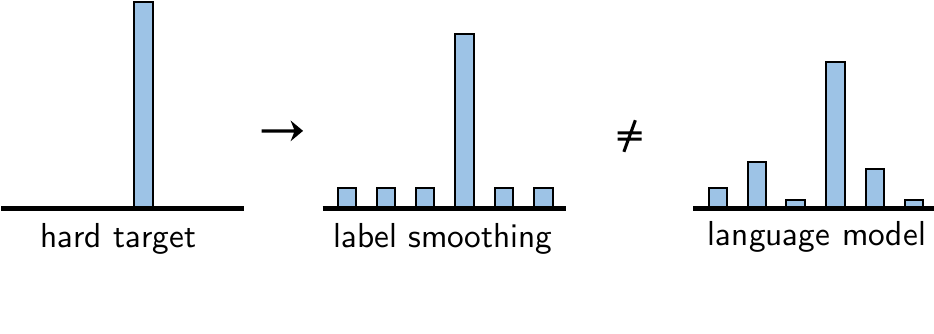}
	\caption{Targets with \ls and \lm-prior.}
	\label{fig:ls}
\end{figure}

\subsection{Relation to Label Smoothing}
Label smoothing (\ls)~\cite{szegedy2016rethinking} is a ``trick'' widely used in machine translation that also uses soft targets.
Specifically, the target distribution at each step is the weighted average between the one-hot distribution $y_k$ of the ground-truth label and a uniform distribution over all other $K$ labels, parameterized by a smoothing parameter $\alpha$: $y_i^{\textsc{ls}} = y_i (1 - \alpha) + \alpha/K$.
The purpose of \ls is to penalize confidence (i.e., low-entropy distributions).

We note that \ls differs from the \lm-prior in two ways.
First, \ls encourages the model to assign \textit{equal} probability to all incorrect words~\cite{muller2019does}, which can be interpreted as a form of uninformative prior (Fig.~\ref{fig:ls}). 
By contrast, the distributions of the \lm are informative, because they express the beliefs of the \lm at each step.
Second, \ls \textit{changes} the target distribution (i.e., first term in Eq.~\eqref{eq:prior}), whereas the \lm-prior involves an additional term, hence the two methods are orthogonal.

\begingroup
\setlength{\tabcolsep}{10pt} %
\renewcommand{\arraystretch}{1.3} %
\begin{table}[tb]
\centering
\small
\begin{tabular}{lrrr}
\hline
\textbf{language-pair}   & \textbf{train} & \textbf{dev} & \textbf{test} \\
\hline
English-Turkish      & 192,482  & 3,007 & 3,000 \\
English-German       & 275,561  & 3,004 & 2,998 \\
\hline
\end{tabular}
\caption{Dataset statistics after preprocessing.}
\label{table:datasets}
\end{table}
\endgroup

\section{Experiments}\label{sec:experiments}
\paragraph{Datasets}
We use two low-resource language pairs (Table~\ref{table:datasets}): the English-German (EN-DE) News Commentary v13 provided by WMT~\cite{bojar-etal-2018-findings}~\footnote{http://www.statmt.org/wmt18/translation-task.html} and the English-Turkish (EN-TR) WMT-2018 parallel data from the SETIMES2\footnote{http://opus.nlpl.eu/SETIMES2.php} corpus.
We use the official WMT-2017 and 2018 test sets as the development and test set, respectively.

As monolingual data for English and German we use the News Crawls 2016 articles~\cite{bojar-etal-2016-findings} and for Turkish we concatenate \textit{all} the available News Crawls data from 2010-2018, which contain 3M sentences.
For English and German we subsample 3M sentences to match the Turkish data, as well as 30M to measure the effect of stronger \lm{s}. 
We remove sentences longer than 50 words. 

\paragraph{Pre-processing} 
We perform punctuation normalization and truecasing and remove pairs, in which either of the sentences has more than 60 words or length ratio over 1.5.
The text is tokenized with sentencepiece~(SPM; \citet{kudo-richardson-2018-sentencepiece}) with the ``unigram'' model.
For each language we learn a separate SPM model with 16K symbols, trained on its respective side of the parallel data.
For English, we train SPM on the concatenation of the English-side of the training data from each dataset, in order to have a single English vocabulary and be able to re-use the same \lm.

\begingroup
\setlength{\tabcolsep}{14pt} %
\renewcommand{\arraystretch}{1.2}
\begin{table}[tb]
	\centering
	\small
    \begin{tabular}{lrr}
    \hline
    \textbf{parameter}               & \multicolumn{2}{c}{\textbf{value}} \\
    \cline{2-3}
                            & \tm          & \lm          \\ 
                            \hline
    Embedding size          & 512         & 1024        \\
    Transformer hidden size & 1024        & 4096        \\
    Transformer layers      & 6           & 6           \\
    Transformer heads       & 8           & 16          \\
    Dropout (all)           & 0.3         & 0.3         \\ 
    \hline
    \end{tabular}
	\caption{Hyperparameters of the \tm{s} and \lm{s}.}
	\label{table:hyperparams-transformer}
\end{table}
\endgroup

\begingroup
\setlength{\tabcolsep}{16pt} %
\renewcommand{\arraystretch}{1.2} %
\begin{table}[tb]
	\centering
	\small
	\begin{tabular}{lrr}
		\hline
		\textbf{language}  & \textbf{3M} (\small{\textsc{ppl$\downarrow$}})   & \textbf{30M} (\small{\textsc{ppl$\downarrow$}})   \\
		\hline
		English   & 29.70          & \textbf{25.02} \\ 
		\noalign{\vskip 2pt}    
		German   & 22.71          & \textbf{19.22} \\
		\noalign{\vskip 2pt} 
		Turkish  & \textbf{22.78} & --             \\
		\hline
	\end{tabular}
	\caption{Perplexity scores for \lm{s} trained on each language's monolingual data, computed on a small held-out validation set per language.}
	\label{table:lm-short}
\end{table}
\endgroup
\paragraph{Model Configuration}
In all experiments, we use the Transformer architecture for both the \lm{s} and \tm{s}.
Table~\ref{table:hyperparams-transformer} lists all their hyperparameters.
For the \tm{s} we found that constraining their capacity and applying strong regularization was crucial, otherwise they suffered from over-fitting.
We also found that initializing all weights with \textit{glorot-uniform}~\cite{pmlr-v9-glorot10a} initialization 
and using pre-norm residual connections~\cite{xiong2020on, nguyen2019transformers}, improved stability.
We also tied the embedding and the output (projection) layers of the decoders~\cite{E17-2025, Inan2017TyingWV}. 

We optimized our models with Adam~\cite{kingma2014Adam} with a learning rate of 0.0002 and a linear warmup for the first 8K steps, followed by inverted squared decay and with mini-batches with 5000 tokens per batch.
We evaluated each model on the dev set every 5000 batches, by decoding using greedy sampling, and stopped training if the \bleu score did not increase after 10 iterations.

For the \lm training we followed the same optimization process as for the \tm{s}. 
However, we use Transformer-large configuration, in order to obtain a powerful \lm-prior. 
Crucially, we did not apply \ls during the \lm pretraining, because, as discussed, it pushes the models to assign equal probability to all incorrect words~\cite{muller2019does}, which will make the prior less informative. 
In Table~\ref{table:lm-short} we report the perplexities achieved by each \lm on different scales of monolingual data.

We developed our models in PyTorch~\cite{NEURIPS2019_9015} and we used the Transformer implementation from JoeyNMT~\cite{JoeyNMT}.
We make our code publically available\footnote{\url{github.com/cbaziotis/lm-prior-for-nmt}}.

\newcommand{\perf}[2]{$\text{#1}\scriptscriptstyle\pm\text{#2}$}
\newcommand{\best}[2]{$\textbf{#1}\scriptscriptstyle\pm\text{#2}$}
\newcommand{\gain}[2]{$\text{\ul{#1}}\scriptscriptstyle\pm\text{#2}$}
\newcommand{\trans}[2]{$\textbf{#1\tto{#2}}$}
\begingroup
\setlength{\tabcolsep}{9.0pt} %
\renewcommand{\arraystretch}{1.3} %
\setuldepth{0}
\begin{table*}[t]
	\small
\begin{tabular}{lcccccccc}
	\hline
	\multirow{2}{*}{Method} & \multicolumn{2}{c}{\trans{DE}{EN}}   &\multicolumn{2}{c}{\trans{EN}{DE}} &\multicolumn{2}{c}{\trans{TR}{EN}}    &\multicolumn{2}{c}{\trans{EN}{TR}}  \\
	\cline{2-9}
	                        & \textbf{dev}     & \textbf{test}    & \textbf{dev}     & \textbf{test}    & \textbf{dev}     & \textbf{test}    & \textbf{dev}     & \textbf{test}    \\
	\hline
	Base                    & \perf{22.6}{0.1} & \perf{26.9}{0.1} & \perf{18.3}{0.3} & \perf{25.6}{0.2} & \perf{15.9}{0.0} & \perf{16.6}{0.3} & \perf{12.2}{0.1} & \perf{11.2}{0.2} \\
	Shallow-fusion          & \perf{23.4}{0.1} & \perf{27.8}{0.1} & \perf{18.5}{0.2} & \perf{26.0}{0.1} & \perf{16.5}{0.1} & \perf{17.3}{0.3} & \perf{12.7}{0.0} & \perf{11.5}{0.1} \\
	\textsc{postnorm}       & \perf{20.4}{0.2} & \perf{24.5}{0.3} & \perf{16.6}{0.1} & \perf{22.9}{0.3} & \perf{13.8}{0.2} & \perf{14.8}{0.1} & \perf{11.0}{0.1} & \perf{10.2}{0.2} \\
	\textsc{postnorm}+ \ls & \perf{22.0}{0.3} & \perf{26.4}{0.2} & \perf{16.9}{0.5} & \perf{23.3}{0.5} & \perf{15.0}{0.1} & \perf{16.0}{0.0} & \perf{12.5}{0.2} & \perf{11.0}{0.2} \\
	Base + \ls              & \perf{23.8}{0.6} & \perf{28.4}{0.7} & \perf{19.2}{0.3} & \perf{27.3}{0.3} & \perf{17.5}{0.1} & \perf{18.4}{0.2} & \perf{13.8}{0.2} & \perf{12.6}{0.0} \\
	Base + Prior       & \best{24.9}{0.0} & \best{30.2}{0.1} & \best{20.5}{0.3} & \best{29.1}{0.7} & \best{18.5}{0.2} & \best{19.5}{0.2} & \best{15.1}{0.1} & \best{13.8}{0.1} \\
	\hline
	Base + Prior + \ls & \gain{25.1}{0.3} & \gain{30.3}{0.3} & \gain{20.8}{0.4} & \gain{29.7}{0.7} & \perf{18.5}{0.3} & \perf{19.5}{0.2} & \gain{15.5}{0.1} & \gain{14.1}{0.2} \\
	Base + Prior (30M)      & \perf{24.9}{0.1} & \perf{30.0}{0.1} & \gain{21.0}{0.4} & \gain{29.8}{0.3} & \gain{18.6}{0.0} & \perf{19.5}{0.2} & --               & --               \\
	\hline
\end{tabular}
	\caption{\bleu scores of each model. Mean and stdev of 3 runs reported. The top section contains the main results, where all methods use \lm{s} trained on the \textit{same} amount of data (3M). The bottom section compares different configurations of the \lm-prior. \ul{Underlined} scores denote gains over the ``Base + Prior (3M)'' model.}
	\label{table:results}
\end{table*}
\endgroup

\subsection{Experiments}
We compare the proposed \lm-prior with other approaches that incorporate a pretrained \lm or regularize the outputs of the \tm.
First, we consider a vanilla \nmt baseline without \ls.
Next, we compare with fusion techniques, namely \textit{shallow-fusion}~\cite{gulcehre2015using} and \textsc{postnorm}~\cite{stahlberg-etal-2018-simple}, which in the original paper outperformed other fusion methods. 
We also separately compare with label smoothing (\textsc{ls}), because it is another regularization method that uses soft targets.
We report detokenized case-sensitive \bleu using sacre-\bleu~\cite{post-2018-call}\footnote{Signature ``BLEU+c.mixed+\#.1+s.exp+tok.13a+v.1.4.2''}, and decode with beam search of size 5.
The \lm{s} are \textit{fixed} during training for both \textsc{postnorm} and the prior.

We tune the hyper-parameters of each method on the DE\tto{EN} dev-set.
We set the interpolation weight for \textit{shallow-fusion} to $\beta$=0.1, the smoothing parameter for \textsc{ls} to $\alpha=0.1$. For the \lm-prior we set the regularization weight to $\lambda$=0.5 and the temperature for $\mathcal{L}_{\textsc{kl}}$ to $\tau$=2.

\subsection{Results}\label{sec:results}
First, we use in all methods \lm{s} trained on the \textit{same} amount of monolingual data, which is 3M sentences.
We used the total amount of available Turkish monolingual data (3M) as the lowest common denominator.
This is done to remove the effects of the size of monolingual data from the final performance of each method, across language-pairs and translation directions.
The results are shown in the top section of Table~\ref{table:results}.
We also report results with recurrent neural networks (\rnn) based on the attentional encoder-decoder~\cite{Bahdanau2014} architecture in appendix~\ref{sec:appendix}.

Overall, adding the \lm-prior consistently improves performance in all experiments. Specifically, it yields up to +1.8 \bleu score gains over the strongest baseline ``Base+\textsc{ls}'' (\textsc{de}\tto{\textsc{en}} and \textsc{en}\tto{\textsc{de}}).
This shows that the proposed approach yields clear improvements, even with limited monolingual data (3M).
As expected, \textsc{ls} proves to be very effective for mitigating over-fitting in such low-resource conditions.
However, simply penalizing confidence helps up to a point, which is shown by the performance gap between ``Base+\textsc{ls}'' and ``Base+prior''. We explore this further next (\S~\ref{sec:analysis}).

\textit{Shallow-fusion} achieves consistent but marginal improvements in all language-pairs.
It works by making small (local) changes to $p_{\tm}$, which primarily helps improve fluency when the \tm is very unsure about what to generate next.
Surprisingly, when training the \tm with the \textsc{postnorm} objective, it barely reaches the baseline. 
As we show in our analysis (\S~\ref{sec:analysis}), under \textsc{postnorm} the \tm generates very sharp distributions, 
which is a result of how it combines $p_{\tm}$ and $p_{\lm}$\footnote{By multiplying the probabilities of $p_{\tm}$ and $p_{\lm}$, or adding their log-probabilities, very small subset of tokens that have non-negligible probability in \textit{both} of them, will be assigned very large probability in the final distribution}.
We identify two potential reasons for this result. 
First, in ~\cite{stahlberg-etal-2018-simple} \textsc{postnorm} was only tested with \textsc{ls}, which to some extend hid the issue of low-entropy outputs.
To verify this, we trained \textsc{postnorm} with \textsc{ls}. 
We observed that in this case, the scores improve significantly, but it still falls short in comparison with the other methods.
Second, we note that the \lm{s} used in the original paper were also trained with \textsc{ls}. 
We hypothesize that by using an \lm that emitted smoother distribution, it implicitly down-weighted the contribution of $p_{\lm}$, that is similar to the small weight used in \textit{shallow-fusion}, which works better in our experiments.

\begin{figure}[t]
\centering
	\includegraphics[width=1.0\columnwidth]{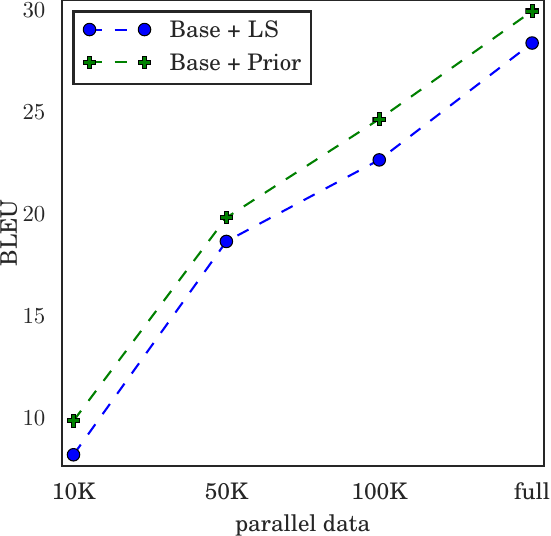}
	\caption{\bleu scores (mean of 3 runs) on the \textsc{de}\tto{\textsc{en}} test set with different scales of parallel data, using the \lm trained on 30M English sentences.}
	\label{fig:efficiency}
\end{figure}

\paragraph{Stronger \lm{s}}
Next, we test how different variations of the \lm-prior affect the translation quality (bottom section of Table~\ref{table:results}). 
First, we lift the monolingual data constraint, in order to evaluate the impact of stronger \lm-priors.
Specifically, for English and German we use \lm{s} trained on 30M sentences.
We observe that the stronger \lm{s} yield improvements only in the \textsc{en}\tto{\textsc{de}} direction.
This could partially be explained by the fact that German has richer morphology than English.
Therefore, it is harder for the decoder to avoid grammatical mistakes in low-resource settings while translating into German, 
and a stronger prior is more helpful for \textsc{x}\tto{\textsc{de}} than \textsc{x}\tto{\textsc{en}}.

However, it is still surprising that the stronger English \lm does not boost performance.
We hypothesize that this might be related to the limited capacity of the \tm{s} we used.
Specifically, in the \kd literature it has been found that the student's performance is affected by the difference between the capacities of the student and teacher networks~\cite{cho2019efficacy, Zhou2020Understanding}. 
In preliminary experiments we also used big \lm{s} pretrained on generic large-scale data, such as \textsc{gpt-2}~\cite{radford2019language}, but we failed to achieve any measurable improvements over the baseline.
Besides the discrepancy in the capacity between the \lm and the \tm, we suspect that another obstacle in this case is the large vocabulary size used in \textsc{gpt-2} (50K symbols).
In particular, \citet{sennrich-zhang-2019-revisiting} showed that in low-resource \nmt, using a very small vocabulary (2K-10K symbols) is the most important factor that affects translation performance.
A potential solution could be to finetune \textsc{gpt-2} on the small vocabulary of the \tm~\cite{zhao2019extreme} and then use it as a prior, but we leave this exploration for future work.

\paragraph{Prior + \textsc{ls}}
We also evaluate a combination of the \lm-prior with \textsc{ls}. 
We observe that in most experiments it has small but additive effects.
This implicitly suggests that the two approaches are complementary to each other.
\ls smooths the one-hot target distribution, which penalizes confidence, whereas the \lm-prior helps improve fluency.
We further explore their differences in our analysis (\S.~\ref{sec:analysis}), by showing the effects each method has on the \tm's distributions.

\subsection{Extremely Low-Resource Experiments}
We also conducted experiments that measure the effect of the \lm-prior on different scales of parallel data.
Specifically, we emulate more low-resource conditions, by training on subsets of the EN\tto{DE} parallel data.
In Fig.~\ref{fig:efficiency} we compare the \bleu scores of the ``Base+\ls'' and the ``Base+Prior (30M)''.

Overall, we observe that adding the \lm-prior yields consistent improvements, even with as little as 10K parallel sentences.
The improvements have a weak correlation with the size of parallel data. 
We hypothesize that by exposing the \tm to a larger sample of target-side sentences, it has the opportunity to extract more information from the prior.
However, we anticipate that in more high-resource settings the improvements will start to diminish.

\begin{figure}[t]
\centering
	\includegraphics[width=1.0\columnwidth]{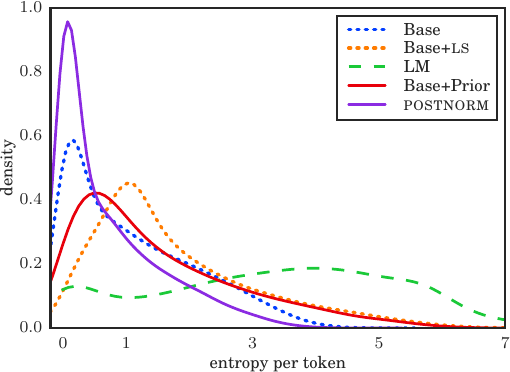}
	\caption{Estimated densities based on each model's entropy on the \textsc{de}\tto \textsc{en} test set.}
	\label{fig:entropy}
\end{figure}

\section{Analysis}\label{sec:analysis}
The main results show that \ls, that simply penalizes confidence, is a very effective form of regularization in low-resource settings. 
We conduct a qualitative comparison to test whether the improvements from the proposed \lm-prior are due to penalizing confidence, similar to \textsc{ls}, or from actually using information from the \lm.
Specifically, we evaluate each model on the \textsc{de}\tto{\textsc{en}} test-set and for each target token we compute the entropy of each model's distribution.
In Fig.~\ref{fig:entropy} we plot for each model the density\footnote{We fit a Gaussian kernel with bandwidth 0.3 on the entropy values of each model. Density plots are more readable compared to plotting overlapping histograms.}
over all its entropy values.

First, we observe that the un-regularized ``Base'' model generates very confident (low-entropy) distributions, which suggests that it overfits on the small parallel data. As expected, the \textsc{ls} regularization successfully makes the \tm less confident and therefore more robust to over-fitting.
For additional context, we plot the entropy density of the \lm and observe that, unsurprisingly, it is the most uncertain, since it is unconditional.

Interestingly, the model trained with the \lm-prior emits \textit{more} confident distributions than the ``Base+\textsc{ls}'' model, although it also achieves significantly better performance.
This clearly shows that the gains cannot be explained just from smoothing the distributions of the \tm and suggests that the model indeed exploits information from the \lm.

\begin{figure}[t]
	\includegraphics[width=1.0\columnwidth]{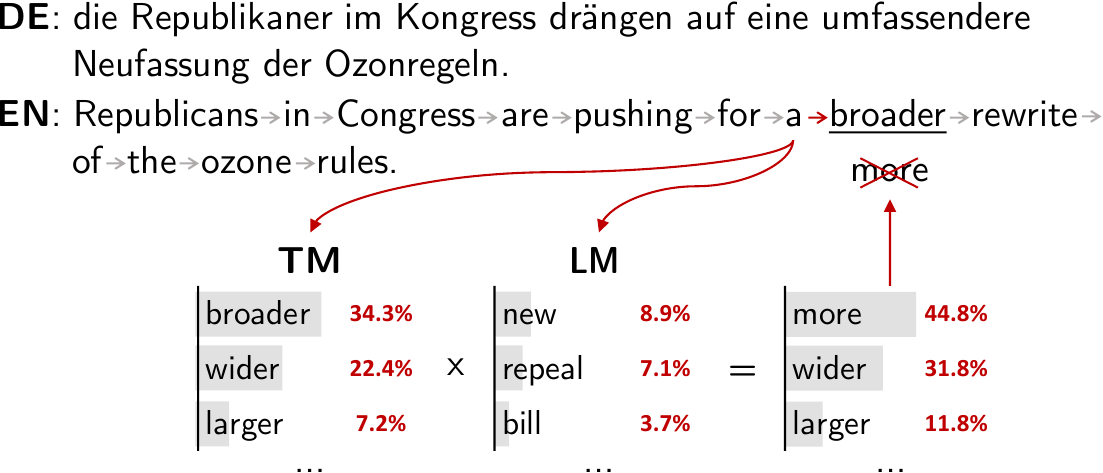}
	\caption{Example of failure of probability interpolation between \lm and \tm, while translating \textsc{de}\tto\textsc{en}.}
	\label{fig:postnorm}
\end{figure}

Next, we focus on the ``Base+\textsc{postnorm}'' model and observe that it generates the most confident predictions.
Note that, this finding aligns with a similar analysis in the original paper, where it was shown that under \textsc{postnorm} the \tm generates low-entropy distributions.
However, even though this method might improve fluency, it can hurt translation quality in certain cases.
As described in Sec.~\ref{sec:lm-prior}, by multiplying the two distributions, only a small subset of words will have non-zero probability in the final distribution.
This means that when there are ``disagreements'' between the \tm and \lm this can lead to wrong predictions.
We illustrate this with a concrete example in Fig.~\ref{fig:postnorm}. 
Although the \tm predicted the correct word, the multiplication with the \lm distribution caused the model to finally make a wrong prediction.
Also, the final distribution assigns a relatively high probability to a word (``more''), which is not among the top predictions of neither the \lm or the \tm.
By contrast, the \lm-prior does \textit{not} change the \tm's predictions, and the model has the flexibility to deviate from the prior.

\begin{figure}[b]
\centering
	\includegraphics[width=1.0\columnwidth]{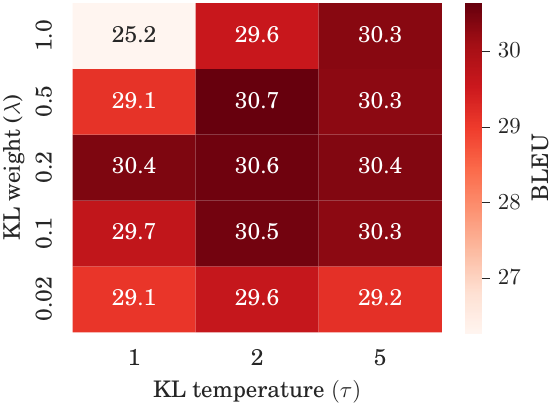}
	\caption{\bleu scores on the DE-EN dev set of models trained with different $\lambda$ and $\tau$ for the $\mathcal{L}_{\textsc{kl}}$.
	Mean of 3 runs for each combination reported.}
	\label{fig:sensitivity}
\end{figure}

\subsection{$\mathcal{L}_{\textsc{kl}}$ Sensitivity Analysis}
The proposed regularization uses two different hyper-parameters in $\mathcal{L}_{\textsc{kl}}$, the weight $\lambda$ that controls the strength of the regularization, and the temperature $\tau$ that controls how much information from the long-tail of the \lm to expose to the \tm. 
We do a pairwise comparison between them, in order to measure how sensitive the model is to their values. 
In Fig.~\ref{fig:sensitivity} we plot a heatmap of the \bleu scores achieved by models trained on the DE\tto{EN} dev-set with various combinations.

Overall, we observe a clear pattern emerging of how the \lm-prior affects performance, which suggests that (1) using  $\tau>1$ indeed helps the \tm to acquire more of the knowledge encoded in the prior, and (2) increasing the strength of the regularization up to a point yields consistent improvements. 
We find that the performance is less sensitive to the value of $\tau$, compared to $\lambda$ and that by setting $\tau>1$, the model becomes also more robust to $\lambda$.
Our explanation is that for $\tau>1$, the \tm tries to match a larger part of the \lm'{s} distribution and focuses less on its top-scoring words.
Therefore, it is reasonable to observe that in the extreme case when we set equal weight to the $\mathcal{L}_{\textsc{mt}}$ and $\mathcal{L}_{\textsc{kl}}$ ($\lambda=1$) and $\tau=1$ the performance starts to degrade, because we strongly push the \tm to match only the top-scoring predictions of the \lm, that is unconditional.
This forces the \tm to pay less attention to the source sentence, which leads to translation errors.

\section{Related Work}
Most recent related work considers large pretrained models, either via transfer-learning or feature-fusion. 
\citet{Zhu2019-su, Clinchant2019-ai, Imamura2019-fo} explore combinations of using BERT as initialization for \nmt, or adding BERT's representations as extra features.
\citet{Yang2019-rc} address the problem of \textit{catastrophic-forgetting} while transferring BERT in high-resource settings, with a sophisticated fine-tuning approach.
In concurrent work, \citet{chen2019distilling} propose knowledge-distillation using BERT for various text generation tasks, including \nmt, by incentivizing the sequence-to-sequence models to ``look into the future''. 
However, in our work we address a different problem (low-resource \nmt) and have different motivation.
Also, we consider auto-regressive \lm{s} as priors, which have clear interpretation, unlike BERT that is not strictly a \lm and requires bidirectional context. 
Note that, large pretrained \lm{s}, such as BERT or \textsc{gpt-2}, have not yet achieved the transformative results in \nmt that we observe in natural language understanding tasks (e.g., GLUE benchmark~\cite{wang2018glue}).

There are also other approaches that have used posterior regularization to incorporate prior knowledge into \nmt.
\citet{zhang-etal-2017-prior} exploit linguistic real-valued features, such as dictionaries or length ratios, to construct the  distribution for regularizing the \tm's posteriors.
Recently, \citet{ren2019unsupervised} used posterior regularization for unsupervised \nmt, by employing an \smt model, which is robust to noisy data, as a prior over a neural \tm to guide it in the iterative back-translation process.
Finally, \lm{s} have been used in a similar fashion as priors over latent text sequences in discrete latent variable models~\cite{miao2016,havrylov2017,baziotis-etal-2019-seq}.

\section{Conclusions}

In this work, we present a simple approach for incorporating knowledge from monolingual data to \nmt.
Specifically, we use a \lm trained on target-side monolingual data, to regularize the output distributions of a \tm.
This method is more efficient than alternative approaches that used pretrained \lm{s}, because it is not required during inference.
Also, we avoid the translation errors introduced by \lm-fusion, because the \tm is able to deviate from the prior when needed.

We empirically show that while this method works by simply changing the training objective, it achieves better results than alternative \lm-fusion techniques.
Also, it yields consistent performance gains even with modest monolingual data (3M sentences) across all translation directions.
This makes it useful for low-resource languages, where not only parallel but also monolingual data are scarce.

In future work, we intend to experiment with the \lm-prior under more challenging conditions, such as when there is domain discrepancy between the parallel and monolingual data.
Also, we would like to explore how to overcome the obstacles that prevent us from fully exploiting large pretrained \lm{s} (e.g., GPT-2) in low-resource settings.

\section*{Acknowledgments}
\lettrine[image=true, lines=2, findent=1ex, nindent=0ex, loversize=.15]{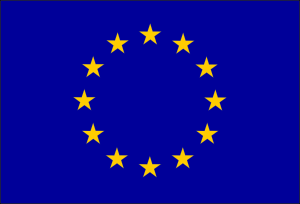}%
{T}his work was conducted within the scope of the \ProjectType\ \textit{\ProjectName}, which has received funding from the European Union's Horizon 2020 research and innovation programme under grant agreement No \GrantNo. 

It was also supported by the UK Engineering and Physical Sciences Research Council fellowship grant EP/S001271/1 (MTStretch).

It was performed using resources provided by the Cambridge Service for Data Driven Discovery (CSD3) operated by the University of Cambridge Research Computing Service (\url{http://www.csd3.cam.ac.uk/}), provided by Dell EMC and Intel using Tier-2 funding from the Engineering and Physical Sciences Research Council (capital grant EP/P020259/1), and DiRAC funding from the Science and Technology Facilities Council (\url{www.dirac.ac.uk}).

\bibliography{refs}
\bibliographystyle{acl_natbib}

\clearpage
\newpage

\clearpage
\appendix
\section{Appendix}\label{sec:appendix}
Our preliminary experiments were conducted with recurrent neural networks (\rnn), because we faced stability problems with the Transformer-based \tm{s}.
We include those results here for completeness.
The experiments were conducted with the 3M monolingual data in all translation directions, therefore they match the top section of the main results reported in the paper (Table~\ref{table:results}).
We observe the same relative performance as with the Transformer-based models, which verifies that the proposed approach transfers across architectures.
However, the differences are smaller, because the \rnn-based models achieved overall worse \bleu scores and perplexities, for the translation and language modeling tasks, respectively.

\begingroup
\setlength{\tabcolsep}{14pt} %
\renewcommand{\arraystretch}{1.3}
\begin{table}[b]
	\centering
	\small
	\begin{tabular}{lrr}
		\hline
		\textbf{parameter}               & \multicolumn{2}{c}{\textbf{value}} \\
		\cline{2-3}
		                        & \tm    & \lm  \\ 
		\hline
		Embedding size (all)    & 256    & 512 \\
		Embedding dropout (all) & 0.2    & 0.2  \\
		Encoder size            & 512    & --   \\
		Encoder layers          & 2      & --   \\
		Encoder dropout         & 0.2    & --   \\
		Decoder size            & 512    & 1024 \\
		Decoder layers          & 2      & 2    \\
		Decoder dropout         & 0.2    & 0.2  \\
		Attention function      & global & --   \\
		\hline
	\end{tabular}
	\caption{Hyperparameters of \rnn-based \tm{s} and \lm{s}.}
	\label{table:hyperparams}
\end{table}
\endgroup
\paragraph{Model Configuration}\label{sec:appendix-model}
We employ the attentional encoder-decoder~\cite{Bahdanau2014} architecture, using the ``global'' attention mechanism~\cite{luong2015}.
The recurrent cells are implemented using Long short-term memory (LSTM;~\citet{hochreiter1997lstm}) units. 
We use a bidirectional \textsc{lstm} encoder and a unidirectional \textsc{lstm} decoder. 
We also tie the embedding and the output (projection) layers of the decoders~\cite{E17-2025, Inan2017TyingWV}. 
and apply layer normalization~\cite{ba2016layer} to the last decoder representation, before the softmax. 

We did not do any hyperparameter tuning, but selected the hyper-parameter values based on~\citet{sennrich-zhang-2019-revisiting}, while also trying to keep approximately the same number of parameters as their Transformer-based counterparts.
Table~\ref{table:hyperparams} lists all the model hyperparameters.
All models were optimized with the Adam optimizer~\cite{kingma2014Adam} with a learning rate of 0.0002 and with mini-batches with 2000 tokens per batch. 

\paragraph{Language Models}
For the \lm{s} we used an identical architecture as the decoder of the \tm, but with larger size.
We also followed the same optimization process. Table~\ref{table:hyperparams} lists all the \rnn-\lm hyperparameters.

\subsection{Language Models}
For completeness, we include here some additional information about the training of the \lm{s}.
In all experiments we paired the \tm with \lm{s} with the same architecture, in order to evaluate how the proposed approach generalizes across architectures.
We train one \lm for each language, on its respective monolingual corpus.
For the Transformer-based \lm{s} we also used a larger corpus for the high resource languages, as a part of our comparison shown in the main body of the paper. 
To evaluate the performance of the \lm{s} and to perform early stopping we used a small held-out development set with 10K sentences.
Specifically, we stopped training when the perplexity on the development was not improved on for more than 10 epochs.
In Table~\ref{table:lm} we report the perplexities achieved by the \lm{s} on each monolingual corpus.
\begingroup
\setlength{\tabcolsep}{8pt} %
\renewcommand{\arraystretch}{1.2} %
\begin{table}[htbp]
	\centering
	\small
	\begin{tabular}{llrr}
		\hline
		\textbf{language} & \textbf{model}     &\multicolumn{2}{l}{\textbf{sentences} (\small{\textsc{ppl$\downarrow$}})}\\
		\cline{3-4} 
		        &                    & \textbf{3M}    & \textbf{30M}   \\
		\hline
		English & LSTM               & 37.04          & --             \\
		        & Transformer (big)  & 29.70          & \textbf{25.02} \\ 
		\hline
		\noalign{\vskip 2pt}    
		German  & LSTM               & 31.26          & --             \\
		        & Transformer (big)  & 22.71          & \textbf{19.22} \\
		\hline
		\noalign{\vskip 2pt} 
		Turkish & LSTM               & 31.26          & --             \\
		        & Transformer (big)  & \textbf{22.78} & --             \\
		\hline
	\end{tabular}
	\caption{Perplexity (\textsc{ppl $\downarrow$}) scores for \lm{s} trained on each language's monolingual data, computed on a small held-out validation set per language.}
	\label{table:lm}
\end{table}
\endgroup

\newcommand{\num}[2]{$\text{#1}\scriptscriptstyle\pm\text{\tiny#2}$}
\begingroup
\setlength{\tabcolsep}{8.pt} %
\renewcommand{\arraystretch}{1.4} %
\setuldepth{0}
\begin{table*}[htbp]
	\small
	\begin{tabular}{lcccccccc}
		\hline
		\multirow{2}{*}{Method} & \multicolumn{2}{c}{\textbf{DE\tto EN}}                     &\multicolumn{2}{c}{\textbf{EN\tto DE}}                        &\multicolumn{2}{c}{\textbf{TR\tto EN}}                        &\multicolumn{2}{c}{\textbf{EN\tto TR}}\\
		\cline{2-9}
		                  & \textbf{dev}             & \textbf{test}            & \textbf{dev}             & \textbf{test}             & \textbf{dev}             & \textbf{test}            & \textbf{dev}             & \textbf{test}            \\
		\hline
		Base              & \num{19.8}{0.1}          & \num{24.2}{0.2}          & \num{15.9}{0.3}          & \num{21.7}{0.4}           & \num{13.1}{0.1}          & \num{13.4}{0.4}          & \num{\,9.9}{0.1}         & \num{\,9.3}{0.1}         \\
		Shallow-fusion    & \num{20.3}{0.1}          & \num{24.9}{0.3}          & \num{16.0}{0.5}          & \num{22.1}{0.6}           & \num{13.5}{0.2}          & \num{13.8}{0.5}          & \num{10.2}{0.2}          & \num{\,9.7}{0.1}         \\
		\textsc{postnorm} & \num{19.7}{0.2}          & \num{24.0}{0.3}          & \num{15.6}{0.1}          & \num{21.0}{0.3}           & \num{11.9}{0.1}          & \num{12.6}{0.3}          & \num{\,9.8}{0.3}         & \num{\,8.8}{0.2}         \\
		Base + LS         & \num{20.6}{0.1}          & \num{25.2}{0.3}          & \num{16.2}{0.3}          & \num{22.8}{0.2}           & \num{13.7}{0.1}          & \num{14.4}{0.1}          & \num{\ul{10.6}}{0.1}     & \num{\,9.8}{0.2}         \\
		Base + Prior      & \num{\ul{20.7}}{0.3}     & \num{\ul{25.3}}{0.4}     & \num{\ul{16.5}}{0.4}     & \num{\ul{23.0}}{0.7}      & \num{\ul{13.9}}{0.1}     & \num{\ul{14.5}}{0.2}     & \num{10.3}{0.2}          & \num{\ul{\,9.8}}{0.1}    \\
		\hline
		Base + Prior + LS & \num{\textbf{20.8}}{0.2} & \num{\textbf{25.3}}{0.3} & \num{\textbf{16.9}}{0.3} & \num{\textbf{23.53}}{0.2} & \num{\textbf{14.2}}{0.2} & \num{\textbf{14.8}}{0.1} & \num{\textbf{10.6}}{0.2} & \num{\textbf{10.0}}{0.2} \\
		\hline
	\end{tabular}
	\caption{\bleu scores of each \rnn-\nmt method. Mean and standard deviation of 3 runs reported. }
	\label{table:rnn-results}
\end{table*}
\endgroup

\end{document}